\documentclass[letterpaper, 10 pt, conference]{ieeeconf}  

\IEEEoverridecommandlockouts                              

\usepackage{graphics} 
\usepackage{epsfig} 
\usepackage{amsmath} 
\usepackage{amssymb}  
\usepackage{amsfonts}
\usepackage{algorithm,algpseudocode}
\usepackage{cite}
\usepackage{subcaption}
\usepackage{graphicx}
\usepackage{adjustbox}

\algnewcommand\algorithmicforeach{\textbf{for each}}
\algdef{S}[FOR]{ForEach}[1]{\algorithmicforeach\ #1\ \algorithmicdo}
\let\oldReturn\Return
\renewcommand{\Return}{\State\oldReturn}

\title{\LARGE \bf
Real-Time Joint Simulation of LiDAR Perception and Motion Planning for Automated Driving 
}

\author{Zhanhong Huang, Xiao Zhang and Xinming Huang
\thanks{*This work was supported by the US NSF Grant 2006738 and by The MathWorks. The authors are with the Department of Electrical \& Computer Engineering and Department of Computer Science, Worcester Polytechnic Institute, Worcester, MA 01609, USA. {\tt\small \{zhuang5,xzhang25,xhuang\}@wpi.edu }}%
}

\begin{document}
\maketitle
\thispagestyle{empty}
\pagestyle{empty}

\begin{abstract}
Real-time perception and motion planning are two crucial tasks for autonomous driving. While there are many research works focused on improving the performance of perception and motion planning individually, it is still not clear how a perception error may adversely impact the motion planning results. In this work, we propose a joint simulation framework with LiDAR-based perception and motion planning for real-time automated driving. Taking the sensor input from the CARLA simulator with additive noise, a LiDAR perception system is designed to detect and track all surrounding vehicles and to provide precise orientation and velocity information. Next, we introduce a new collision bound representation that relaxes the communication cost between the perception module and the motion planner. A novel collision checking algorithm is implemented using line intersection checking that is more efficient for long distance range in comparing to the traditional method of occupancy grid. We evaluate the joint simulation framework in CARLA for urban driving scenarios. Experiments show that our proposed automated driving system can execute at 25 Hz, which meets the real-time requirement. The LiDAR perception system has high accuracy within 20 meters when evaluated with the ground truth. The motion planning results in consistent safe distance keeping when tested in CARLA urban driving scenarios.
\end{abstract}

\section{INTRODUCTION}
\subsection{Motivation}
Perception and Motion Planning are two crucial tasks in an automated driving system. The perception module in an autonomous vehicle (AV) receives the data from the on-board sensors, process the information instantaneously, and send the vehicle perception results to the motion planner. The task of autonomous driving perception has been studied extensively using different approaches. The traditional object based methods include object detection, object classification\, and object tracking\cite{Geiger2012CVPR}. With advance of deep learning models, frame based methods such as  semantic segmentation\cite{behley2019iccv}, panoptic segmentation\cite{behley2020benchmark}, and 4D panoptic segmentation\cite{ayguen2021cvpr} have become popular. The critical requirements of a perception system are real-time processing with high accuracy and computational efficiency. 
\par Upon receiving the information from the perception module, motion planner makes behavior decisions in real-time, including path planning and velocity planning, and then send commands to the executor, i.e., vehicle's kinematic controller and electronic control unit (ECU). Motion planner must be robust and efficient. In order to meet the strict real-time requirement, it is necessary to build a system in which the perception and motion planner modules communicates and coordinates accordingly. 

\begin{figure}[]
\centering
\includegraphics[width=8cm]{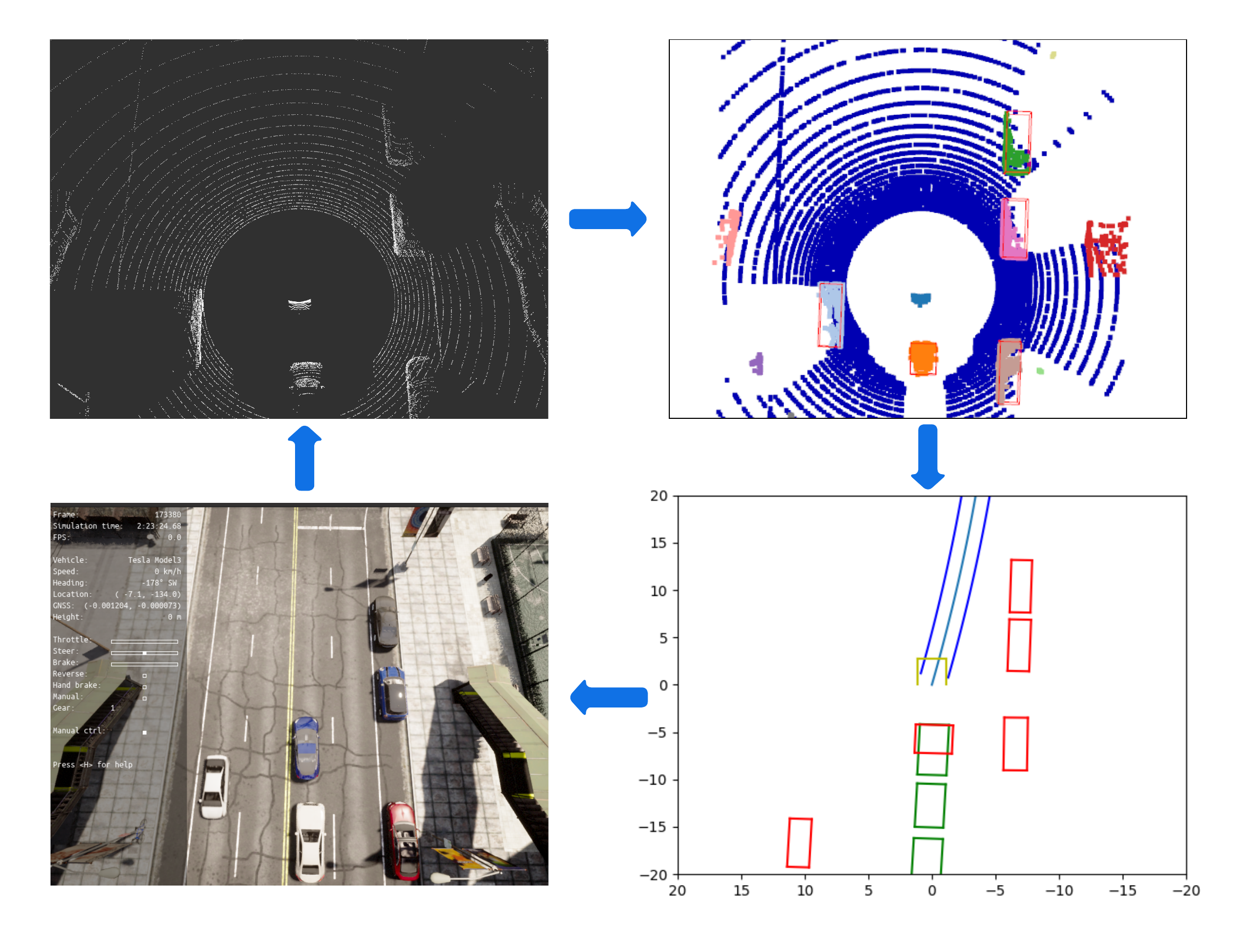}
\caption{Illustration of our system pipeline. Bottom left: CARLA simulator\cite{dosovitskiy_carla_2017}, Top Left: Lidar raw data shown in Rviz. Top Right: visualization of the Lidar perception. Bottom Right: the visualization of the Motion Planning}
\label{fig:system_demo}
\end{figure}

\par The existing works of perception were mainly focused on accuracy. Datasets with metrics are defined as benchmarks for various perception tasks. For object detection, a perception system must segment the objects and predict the bounding box that aligns with the ground truth. Semantic segmentation and panoptic segmentation utilize mIoU to evaluate the performance of point-wise prediction results. Many perception research were aimed at improving the benchmark metrics by a few percentages or even a fraction of one percent. The question is - Does these progressive improvement on the perception benchmark metrics affect the actual automated driving outcomes? In other words, does the small increment in perception performance achieved by expensive computations and energy cost really benefit the motion planning? This is the motivation of our research work.
\par For trajectory planning, collision checking is often a time-consuming yet indispensable step. A general approach in existing works requires discrete representation of the environment in Frenet coordinate system\cite{zhang_hybrid_2018,zheng_bezier_2020,xu_autonomous_2021,jie_real-time_2022}. It first uses the bounding circles to represent the collision constraint of all vehicles and obstacles in the surrounding area on the map. Next, it transforms the ego vehicle along the path on the map to check whether the obstacles would appear in the ego vehicle's bounding representation. The advantage of this method is that the data structure of grid space representation is convenient for path searching. Similar data structure can be directly applied to the sensor input as a geometric representation of all objects on the road. However, this method also has several disadvantages: Firstly, the cost of grid representation is high which is directly proportional to the map resolution and the sensing distance. Because of the high cost of mapping the environment, areas outside of the driving lanes are usually neglected. However, objects not on the ego lane, such as pedestrians and misbehaved vehicles at an intersection, must be considered in the path planning process. Secondly, the traditional method assumes that the geometric representation of the obstacles all appears at once. But in practice the shape of an object may not reveal completely in a LiDAR frame because of partial occlusion.  Thirdly, the 2D grid map is good for representation of positions, but not for other measurements such as velocity. It would be time-consuming to convert all types of sensor data into this grid mapping system.
\par Apart from the issues mentioned above, most existing works evaluates each module in an automated driving system independently. For instance, the path planning module always assumes the perception of the environment is perfect and gets feedback instantaneously\cite{xu_opencda_2023,jie_real-time_2022}. The perception module is assumed with a perfect localization, which is a lack of consideration since the two modules are coupled in an AV system and affects each other interactively. Each sub-module is optimized without being evaluating in a complete automated driving framework. Hence, there is a possibility of leading to unexpected behaviors when integrating these modules together during vehicle operations.
\subsection{Contribution}
In contract to the existing works, the proposed framework validates the integration of both perception and motion planning together in a real-time autonomous driving simulation. The main contributions are summarized as follows:
\begin{itemize}
    \item We build a real-time LiDAR perception system using point cloud clustering and tracking-by-detection. An optimized non-iterative L-shaping method is designed to estimate the vehicle position and orientation accurately. Furthermore, we implement the vehicle velocity estimation via dynamic observations based on the UKF-GNN tracking model.
    \item We introduce a novel and efficient collision bound representation that reduces the communication cost between the perception module and the motion planner. The instance based collision checking can work independently without a grid map representation.
    \item We conduct real-time experiments in CARLA to validate the joint simulation approach. The integrated system with LiDAR perception and motion planning meets the urban driving performance requirement.  
\end{itemize}

\section{Related Work}
\subsection{Vehicle Detection and Tracking}
Tracking-by-detection is a popular method for the task of multi-object tracking (MOT). In this section we summarize the non-learning LiDAR-based detection and tracking algorithms.

\textbf{Detection} of all surrounding objects is an important task in LiDAR point cloud processing for data cleaning, candidate instancing,  feature extraction, and object classification. Voxel downsampling is one of the most popular ways to pre-process the Lidar point cloud. The method preserves the spatial distribution of the points while reducing the point size. Ground segmentation is another essential step in pre-processing. RANSAC plan fitting \cite{groundSeg_RANSAC_improve_2022} and slope-based segmentation proposed by \cite{groundSeg_slop} are two of the representative techniques. They describe the ground as a single plane or continuously extended surface in vertical and horizontal directions. Subsequently, a traditional method including instancing, feature extraction, and object classification is often implemented step-by-step. Point cloud clustering is proved to be an efficient way to instance the object candidates in the point set. Distance-based,  density-based, and range image-based methods like Euclidean clustering, DBSCAN \cite{schubert2017dbscan} and depth clustering\cite{bogoslavskyi2016fast} can be selectively applied for Lidar data\cite{zhao2021technical}. The candidates are then processed by the classifier to tag semantic labels of the objects. Through the recent works\cite{Zhang_Xiao_Coifman_Mills_2020} \cite{Sualeh_Kim_2019}, support vector machine (SVM) \cite{Yan_Duckett_Bellotto_2017}, random forest (RF)\cite{Classifier_RF} and simple rule-based method can efficiently classify the point cloud clusters.

\textbf{Tracking} is a step that takes the detected objects, tracking and identifying them as unique samples using the historical information sequence. This is done through a framework that combines object state prediction (filter) and data association to deal with object dynamics. The extended Kalman filter (EKF) and Unscented Kalman Filter (UKF) are commonly used nonlinear approaches for filtering, with the UKF outperforming the EKF in predicting heavy nonlinear cases.\cite{tracker_filter_comp20173d}. Other filters, such as interactive multiple model (IMM) filter and particle filter can also be used as alternative solutions. \cite{tracker_Filter}. The Lidar-based vehicle tracking studies typically implemented the global nearest neighbor (GNN) and joint probability distribution association filter (JPDAF) methods for data association. The JPDAF outperforms the GNN in complex cases where multiple measurements are very close \cite{Sualeh_Kim_2019}.

\subsection{Collision Checking in Trajectory Planning}
The existing collision checking methods for automated driving can be categorized into two main categories, i.e. footprint based and potential field based.

\par\textbf{Footprint based methods} use a few circles along the ego vehicle axis to bound the vehicle to create the collision bounding, where the number of circles and and their radii vary in different representations. Zigler and Stiller\cite{ziegler_fast_2010} introduced a fast collision checking method, where they decomposed the vehicle shape into several disks and then converted the disks into a rectangle shape to fit in the grid map.
Zhang et al.\cite{zhang_hybrid_2018} 
proposed a hybrid trajectory planning, where they evaluated the clearance of vehicle footprint in the grid map, which was computed based on the distance between the way point and the obstacle in each specific direction. This clearance allows the path optimizer to find the orientation of the path since the cost increases when the clearance gets shorter.

\par\textbf{Potential field based methods} calculated an ellipse around each object, where the ellipse surrounds the reachable pose in several future time steps provided with the object's current velocity and acceleration. Philipp and Goehring\cite{philipp_analytic_2019} proposed a collision octagon check boundary that facilitates the calculation of collision state probability and collision event probability. The octagon representation simplified the bounding rectangle of two possible colliding vehicles to a point and an octagon, where the octagon's vertex was calculated from the Gaussian state distribution of the obstacle.

\section{Joint Simulation Framework}
\begin{figure*}[ht]
\centering
\includegraphics[width=18cm]{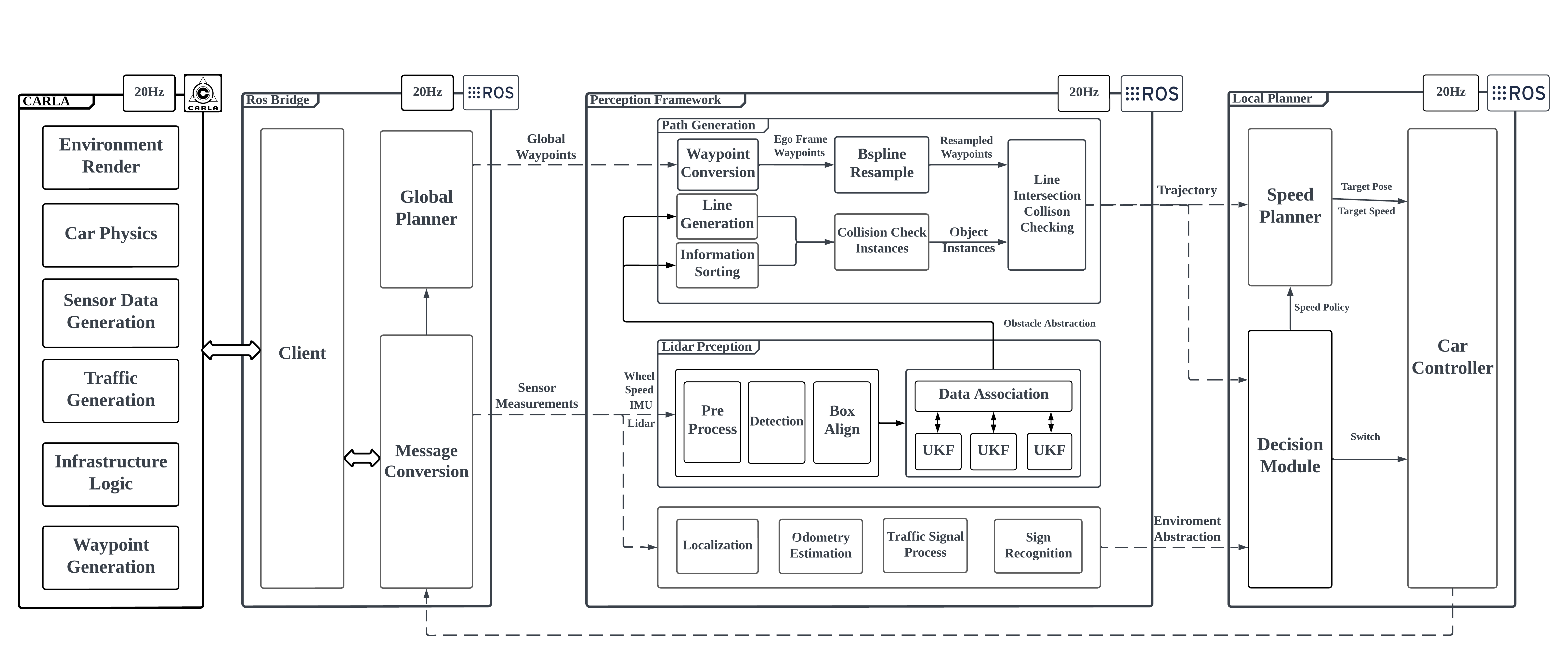}
\caption{Our joint simulation system is based on the CARLA framework and ROS bridge. In the proposed system, the Lidar perception and the motion planner are realized as ROS nodes and work asynchronously. The dashed lines show the asynchronous signals among ROS nodes and the solid lines indicate synchronous signals with each node.}
\label{fig:systemdiagram}
\end{figure*}
Fig. \ref{fig:system_demo} shows our system pipeline and Fig. \ref{fig:systemdiagram} outlines the proposed system framework with all modules and the connections among them. The perception module takes the converted CARLA sensor messages from ROS Bridge as  input\cite{dosovitskiy_carla_2017}. It generates the environment abstractions that allow the decision module in the local planner to decide whether it needs to request global waypoints from the global planner. The module's decision is usually based on the road policies, such as road speed limit changes or ego vehicles stopped by traffic lights. The path generation process starts once the global waypoints are retrieved from the global planner. The path generation module first requests the surrounding obstacles from the Lidar perception module and performs the collision checking. Subsequently, it outputs the collision-free path to the velocity planner for speed estimation and the decision module for loop closure checking. Next, the target pose and speed is sent to the controller to generate the control output. Finally, the control output is sent to the ROS bridge for vehicle movement and display update in CARLA.

\section{Technical Approach}
In this section, we first present the proposed Lidar perception module and its algorithms. Then, we introduce the motion planner module, which includes the collision checking algorithm and the velocity model. 
\subsection{System Assumptions}
For the perception system, we assume the ego car's speed and angular velocity can be fetched by car-deployed sensors such as the wheel tachometer and IMU. Traffic signal status and stop signs can be retrieved from the the image-based sensors or V2X communications. 
For the motion planner, we assume our global waypoints have already been calculated by a remote server, which is not required to be handled by the ego vehicle. We further assume that the ego vehicle gets its global position through GPS, and its position estimation is more accurate than the waypoints resolution.
\subsection{Lidar Perception}

\subsubsection{Pre-processing} As the front-end of a real-time and high-performance Lidar perception system, pre-processing is applied to remove the noise and provide computationally efficient point clouds for downstream processing.

\textbf{Down-sampling.} The time complexity of point cloud algorithms largely depends on the quantity of points. A standard Lidar frame with 64 channels typically consists of approximately 20,000 to 30,000 points, which can be computationally intensive if all points are processed.

Depending on the applications, not all points are equally important. To focus on the potentially moving objects near the ego vehicle, the pre-processing step segments out an origin-centered region of interest (RoI) from a bird-eye view (BEV). Next, a voxel down-sampling method is applied to reduce the negative effects of outliers on the subsequent segmentation and classification processes. 

\textbf{Ground Segmentation.}  As the ground points are naturally connected to nearly all objects in the Lidar scan, distance or density-based point cloud instancing methods do not perform well before ground removal. To address the problem and further reduce the number of points, the RANSAC plane fitting technique is utilized to separate the potential vehicle candidates and the ground plane. Although this iterative RANSAC algorithm is not particularly time efficient, ground plane removal reduces the number of points, which speeds up the subsequent processing considerably.

\subsubsection{Vehicle Detection} We propose an efficient object detection algorithm based on point cloud clustering that does not require deep learning. It effectively extracts the vehicle candidates in each frame.  

\textbf{Vehicle Instancing.} To identify the vehicle candidates from non-ground Lidar points, we implement the DBSCAN method to cluster points based on their spatial density distribution. By configuring the density parameter $\epsilon$ and minimum points within the density region, DBSCAN extracts the clusters with a specific density in a certain range and other sparser distributed points are labeled as noise.

\textbf{L-shape Fitting.}  From our study, the essential requirements of a perception system in local planning for autonomous driving involve determining the position and orientation of the detected objects. To extract the information from each cluster of points, we adopt the L-shape fitting approach. The method fits a rectangle along the L edge from BEV. While L-shape fitting via variance is suitable for 2D point clouds and offers optimal vehicle box alignment, it cannot be directly applied to 3D points due to the large number of points which would require much longer computing time. Instead, we project each cluster of points to BEV and then extract the convex shape of the points. This technique optimizes the run time while preserving the L-shape fitting performance.

\textbf{Rule-based Classifier.} Rule-based classifier is an efficient method for filtering out vehicle candidates based on the geometric and statistical criteria. For vehicle classification, it is highly efficient and provides reliable results when rule criteria are specified accordingly. In this work, we set the thresholds for various parameters such as the number of points, width, length, area, and the width-to-length ratio of the bounding box from BEV for vehicle detection.

\subsection{Multi-Vehicle Tracking}
\textbf{UKF-GNN Framework.} Multi-Vehicle tracking is similar to the classic task of multi-object tracking. The traditional solution comprises two key components: object behavior predictor and data association framework. We implemented the UKF as the predictor while using GNN as the data associator. When the new candidates sequences are fed, the data associator calculates the correlation scores among the objects in the detected list and predicts objects in the tracking list. Based on the score matrix, the tracking list is updated according to the mapping and list management policy. During the update step, the Kalman filter of each object takes the matched detection object information as measurement to update its probability of the states model and subsequently predicts the location and speed of the object for the next state. 

\subsection{Dynamic Velocity Estimation.}
Estimating the velocity of objects from a moving vehicle is considerably more complex than the estimation from stationary observations. This is due to the fact that the current observation coordinates undergo both rotation and translation in the 3D space to its previous observation. Therefore, the initial step in states updating is to adjust the previous observation states using the ego car's speed and angular measurement described by the following equations: 
\begin{equation}\label{eqn:transform}
R(\theta, \Delta x, \Delta y) = \begin{bmatrix}
\cos{\theta} & -\sin{\theta} & \Delta x\\
\sin{\theta} & \cos{\theta} & \Delta y\\
0 & 0 & 1
\end{bmatrix}
\end{equation}

\begin{equation}\label{eqn:ego_travel}
d_{ego} =  v_{ego}\Delta t
\end{equation}

\begin{equation}\label{eqn:ego_current_frame}
\begin{split}
    \Delta x &= - \sin{\theta}d_{ego}\\
    \Delta y &= - \cos{\theta}d_{ego}
\end{split}   
\end{equation}

\begin{equation}\label{eqn:obs_transform}
\begin{bmatrix}
x_{t}\\
y_{t}\\
1
\end{bmatrix} =
R(\theta_{ego}, \Delta x, \Delta y)
\begin{bmatrix}
x_{t-1}\\
y_{t-1}\\
1
\end{bmatrix}
\end{equation}
\newline where $\theta\subseteq[\pi, -\pi)$ is the orientation of ego car in the map coordinate, $\Delta t$ is time difference between each frame. 
For each lidar frame, we have recorded the position of the detected vehicles. Once the object positions in the previous frame are transformed to the current one, we can estimate the velocities and orientations of the tracked vehicles accurately.

\subsection{Motion Planner}

\subsubsection{Waypoint conversion and optimization}
Before applying the perception results for collision checking, we first need to convert our waypoints retrieved from the global path planner into the ego car frame. We consider the path starting from the frame origin, where the origin is set on the ego car's ground center. The frame coordinates are mapped with the right-hand rule, and the \emph{x} positive direction is set to the orientation of the ego car. Given the coordinate frame of the ego car, we can map these waypoints into the ego frame with the transformation matrix:
\begin{equation}\label{eqn:wp_transform}
\begin{bmatrix}
wp_{eg,x}\\
wp_{eg,y}\\
1
\end{bmatrix}
=
R(\theta_{ego}, -p_{x}, -p_{y})
\begin{bmatrix}
wp_{mp,x}\\
wp_{mp,y}\\
1
\end{bmatrix}
\end{equation}
\newline where $\theta_{ego}\subseteq[\pi, -\pi)$ is the orientation of the ego car in the map frame, and $p$ is the position of the ego car on the map. So that we can get a series of waypoints located in the ego frame:
\begin{equation}\label{eqn:waypoint}
    wp = [wp_0, \dots, wp_m]
\end{equation}
where $m$ is the minimum distance for perception, calculated as the safe braking distance based on the ego car's speed:
\begin{equation}\label{eqn:braking_distance}
    d_{brake} = {{v_{ego}}\over {2\mu g}}
\end{equation}
where $\mu$ is the friction coefficient, which we set as a constant equal to 0.35 in our simulation. \emph{g} is the gravitational acceleration, which we also set as a constant equal to 9.8 m/$\text{s}^2$ in our simulation. Given the safe braking distance, we can calculate the number of converted waypoint in each cycle by:
\begin{equation}\label{eqn:num_of_waypoints}
    m = {{d}\over {d_{wp}}}\times f_{safe},\; m \subseteq \mathbf{N}
\end{equation} where the $d_{wp}$ is the preset distance between two waypoints, we applied the safety factor $f_{safe}$ to get an extra safety buffer.
\par Once we get the converted waypoints, we need to use them to calculate a smooth trajectory. In our design, we implement B-Spline to generate the 2-degree Bézier curves, in which the waypoints' \emph{x} and \emph{y} positions are being considered separately, and the reference axis is the travel euclidean distance. Given the reference axis, we can calculate the B-Spline for our trajectory, referring to the x and y axes. Then we can re-sampled the x and y trajectory to get our re-sampled waypoints given a fixed travel distance increment $t_s$, which is set to 0.5 in our simulation:
\begin{equation}\label{eqn:RWpList}
    RWpList = [(trajectory_x(t_s),trajectory_y(t_s)]
\end{equation}

\subsubsection{Collision checking}
Next, our motion planner performs collision checking along the trajectory. In our design, we make an intersection checking between our trajectory and the car bounding boxes. Compared to traditional bubble collision checking, our algorithm is more computationally efficient for long range planning. Moreover, the simplified representation of obstacles can be directly obtained from the lidar perception, which does not need to convert into the BEV. Finally, line intersection checking is more utilized in the detected object motion representation, where the trajectory of the dynamic obstacle and the obstacle itself can be evaluated together.

\begin{figure}[]
\centering
\includegraphics[width=8cm, height=5cm]{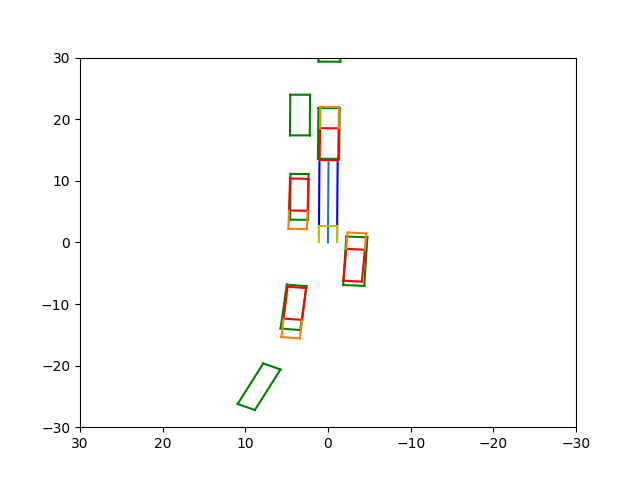}
\caption{An example of the vehicle velocity extended bounding model. The red lines are the lidar detected bounding and the orange lines are the velocity extended bounding. The green lines are the ground truth bounding extended by the estimated velocity.}
\label{fig:extend_demo}
\end{figure}

Based on our assumption that an observed vehicle always drives toward its heading,  we can represent the trajectory of a vehicle by extending the front bounding points, which can be calculated using its speed and orientation:
\begin{equation}\label{eqn:extend}
\begin{split}
    E_x &= \cos{\theta_c}speed_c t_{est}\\
    E_y &= \sin{\theta_c}speed_c t_{est}
\end{split}
\end{equation}

where $theta_c$ is the vehicle orientation in the ego frame, $speed_c$ is the vehicle current speed, and $t_{est}$ is the estimated time for the vehicle trajectory. An example of the vehicle bounding extension model is shown in Fig. \ref{fig:extend_demo}.

By adding the extended distance to the front-end bounding points, we can get an extended bounding box that includes the velocity information in the current time frame. Then we can generate Bounding Lines around the vehicle through the bounding coordinates:
\begin{equation}\label{eqn:BoundingLine}
\begin{split}
    VBound &=[L_1, L_2, L_3, L_4] \\ 
    &=[L_{c1\to c2},L_{c1\to c3}, L_{c2\to c4}, L_{c3\to c4}]
\end{split}
\end{equation}

\begin{figure}[]
\centering
\includegraphics[width=8cm]{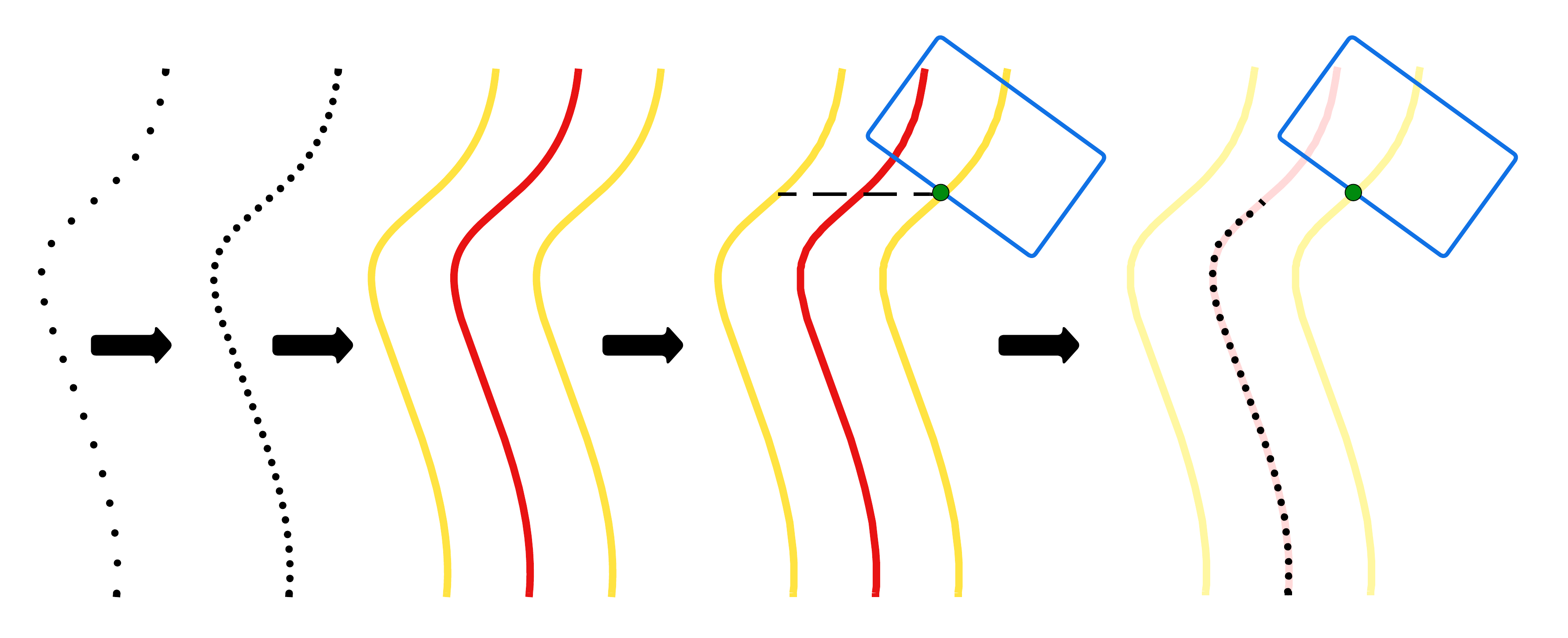}
\caption{Frome left to right: (a) Original global waypoints. (b) Resampled Waypoints after Bspline, trajectory line with side bound assisted check lines (c) Retrieved the interesect point with lowest index. (d) Collision-free path.}
\label{fig:collision_check_alg}
\end{figure}

Next, we sort the $VBound$ along with the information of each vehicle, which is also shown in lines one to six of Algorithm \ref{alg:collision}. Then we perform the intersection check along the trajectory, which is introduced in line seven of Algorithm \ref{alg:collision} and detailed described in Algorithm \ref{alg:intersect} also shown in Fig. \ref{fig:collision_check_alg}. Eventually, after the collision checking, we get the distance, speed, and orientation of the vehicle lying on our trajectory. A new collision-free trajectory is generated.

\begin{algorithm}[]
\caption{Generate Collision Free Path with vehicle info}\label{alg:collision}
\begin{algorithmic}[1]
\Require{$VList, RWpList$}
\Ensure{$distance,speed, \theta, newPath$}
\Statex
\State Init $BoundingList$
\State Init $InfoList$

    \ForEach{$Vehicle$ in $VList$ }
    \newline\Comment{Retrieve vehicle information and bounding box line}
        \State Add List[$bounds$] to $BoundingList$
        \State Add List[$speed, yaw, position$] to $InfoList$
    \EndFor

\Statex
\State $VIndex,distance,newPath\newline\gets\textsc{IntersectCheck}(RWpList, BoundingList) $
\newline\Comment{Get closest vehicle index that occupy on our path, its distance to us and our collision free path}
\Statex
\State$\theta\gets InfoList[VIndex][1]$
\State$speed\gets InfoList[VIndex][2]$
\newline\Comment{Retrieve vehicle measurement from info list}
\Statex
\Return $distance, speed, \theta, newPath$
\end{algorithmic}
\end{algorithm}

\begin{algorithm}[]
\caption{IntersectCheck}\label{alg:intersect}
\begin{algorithmic}[1]
\Require{$RWpList, BoundingList$}
\Ensure{$VIndex, CFWpList$}
\Statex
\State Init $CFWpList\gets RWpList$
    \ForEach{$VBoundSet$ in $BoundingList$}
        \ForEach{$Bound$ in $VBoundSet$}
        \State Init $CheckFlag\gets$ True
        \While{$CheckFlag\;$\textbf{and} $CFWpList$}
        \State $Lines$ generate from $CFWpList$
            \If {$Bound$ Intersect $Lines$}
            \State $CFWpList$.PopRight()
            \State $VIndex\gets$Index($VBoundSet$)
            \State $distance\gets$D($Bound, RWPList[0]$)
            \Else 
            \State $CheckFlag\gets$ False
            \EndIf
        \EndWhile
        \EndFor
    \EndFor
\Return $VIndex$, $distance$, $CFWpList$

\end{algorithmic}
\end{algorithm}

\subsubsection{Speed Planning}
After obtaining the collision-free path and the occupied vehicle information, we need to determine the ego car speed and target pose as the input for the PID controller. In the scenario of lane keeping, given the target pose, speed is calculated for the following three cases:
\begin{equation}\label{eqn:speed_case}
    v_{pre} =
    \begin{cases}
    v_{n} &\text{no obstacle}\\
    v_{obs} &\text{obstacle in front}\\
    v_{platoon} &\text{following vehicle}
    \end{cases}\\
\end{equation}
When there is no obstacle and vehicle in front of the ego car, we calculate the speed by subtracting the estimated reachable distance and the target pose distance multiplied by control timestep $\Delta t$:
\begin{equation}\label{eqn:speed_normal}
    v_{n} =  v_{current}+ (d_{pose}-d_{reach})*\Delta t\\
\end{equation}
When the Intersection check algorithm reports an obstacle in front of the ego car or a vehicle not in the same orientation as the ego car, we cap the speed by estimating the difference between the distance from the obstacle and safe distance $d_{safe}$: 
\begin{equation}\label{eqn:speed_obstacle}
    v_{obs} = v_{appr}+{d_{obs}- d_{safe}\over{d_{safe}}}v_{appr}
\end{equation}\label{eqn:safe_distance}
where $v_{appr}$ is the preset speed for slowing down the ego car. And the safe distance $d_{safe}$ is calculated by $d_{brake}$ in equation \ref{eqn:braking_distance}, and an additional buffer distance $d_{buffer}$ to keep the minimum distance:
\begin{equation}
    d_{safe} = d_{brake}+ d_{buffer}
\end{equation}
When the intersection algorithm finds that there is a leading vehicle in front of the ego car, our target speed is calculated by
\begin{equation}\label{eqn:speed_following}
    v_{platoon} =v_{lead}+ w(d_{lead}-d_{safe})\Delta t
\end{equation}
where $v_{lead}$ is the leading vehicle speed, $d_{lead}$ is the distance between the leading vehicle and the ego car, and $w$ is a proportional weight to apply to the distance difference.
Once the preferred speed is calculated, we bound the speed and send it to the controller.
\begin{equation}\label{eqn:v_exc}
    v_{exc} = 
    \begin{cases}
    0 & v_{pre} < 0\\
    v_{pre} & 0 \leq v_{pre} < v_{reach}\\
    v_{reach} & v_{pre}\geq v_{reach}
    \end{cases}
\end{equation}
where the maximum reachable speed in the next control timestep $v_{reach}$ is calculated with
\begin{equation}\label{eqn:speed_reachable}
    v_{reach} =  v_{current}+ a_{max}*\Delta t,\; v_{reach} \subseteq [v_{init}, v_{max}]
\end{equation}
where $v_{max}$ is the speed limit on the road, and the $v_{init}$ is the initial launch speed that allows the the car ramp up the speed faster from zero.

\section{Empirical Results}
\subsection{Implementation Setup}
We first initialized the CARLA simulator and the CARLA ROS Bridge, which handles the connection between our system and CARLA\cite{dosovitskiy_carla_2017,ros}. We choose Town10 and Town01 as the simulation maps in CARLA with 50 vehicles but no pedestrians. We set a fixed 50 ms as our simulation time-step, ticked by the ROS bridge. For the vehicle speed limit, we set it to 8.33 m/s for Town10, and adaptive max speed based on the speed limit posted on the signs as in Town01. For the $v_{init}$, the initial launch speed, we set it to 6 m/s. The global waypoints were generated by CARLA default waypoint publisher with 2 m resolution. For the trajectory optimization, we re-sampled the trajectory with 0.5 m as the fixed travel distance step $t_s$. For the speed planning,  we set the ego vehicle parameter to $\mu = 0.35$, $a_{max}$ = 2.5 m/$\text{s}^2$. We use the CARLA default PID controller which is already tuned with the ego vehicle as a Tesla Model 3. Each module ran in a ROS1 node and communicated with each other with ROS service protocol. All the experiments were conducted on a desktop computer with an AMD 5950X CPU (4.5GHz).  

\subsection{Motion Planner Verification}
\begin{figure}[]
\centering
\includegraphics[width=8cm, height=3cm]{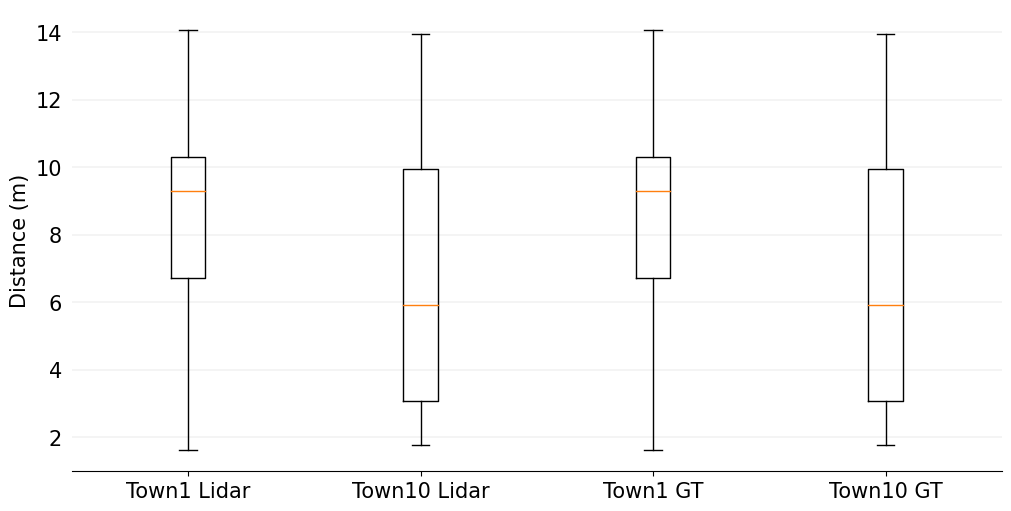}
\caption{Distance to the front vehicle during automated platooning.}
\label{fig:fdistance}
\end{figure}
To validate our motion planner performance, we let the ego vehicle run in both maps with random global waypoints for 40 minutes using ground truth dynamic objects information provided by CARLA. We recorded the distance between ego vehicle and the detected front vehicle in platooning scenarios, which is  shown in the right-hand side of Fig. \ref{fig:fdistance}. We can see that the main distribution of the recorded following distance in Town01 is longer than that of Town10. This is because the speed limit in Town01 is adaptive, which can reach a maximum speed of 15 m/s. For this reason, the closest distance in Town01 is shorter than in Town10. 

\par We also recorded the execution time for path generation and velocity calculation for the reference of system evaluation in the later experiments, shown in the first two columns of Table.\ref{table:time consumption}. We can see the whole system running above 20 Hz. The most time-consuming step is to transform the global waypoints to the ego frame and then back to the global frame after processing. The cases of long processing time occurred when there were too many resampled waypoints to be converted. Because the waypoint number is directly related to the vehicle speed, we can also see the average processing time of Town01 is less than that of Town10.

\begin{table*}[]
\caption{Time consumption for each step}
\label{table:time consumption}
\centering
\begin{tabular}{llccccccccccc}
\hline
                    & \multicolumn{1}{c}{} & \multicolumn{2}{c}{Town01 GT} &  & \multicolumn{2}{c}{Town10 GT} &  & \multicolumn{2}{c}{Town01 Lidar} &  & \multicolumn{2}{c}{Town10 Lidar} \\ \cline{3-4} \cline{6-7} \cline{9-10} \cline{12-13} 
Time                &                      & Avg. (s)      & Max. (s)     &  & Avg. (s)      & Max. (s)      &  & Avg. (s)       & Max. (s)       &  & Avg. (s)        & Max. (s)       \\
Lidar Perception    &                      & /             & /            &  & /             & /             &  & 0.041          & 0.101          &  & 0.034           & 0.090          \\
Path Generation     &                      & 0.009         & 0.045        &  & 0.012         & 0.053         &  & 0.009          & 0.027          &  & 0.011           & 0.024          \\
Speed Planning      &                      & 0.001         & 0.006        &  & 0.002         & 0.03          &  & 0.001          & 0.008          &  & 0.001           & 0.003          \\
Local Planner Total &                      & 0.044         & 0.120        &  & 0.028         & 0.076         &  & 0.050          & 0.143          &  & 0.031           & 0.092          \\ \hline
\end{tabular}
\end{table*}

\subsection{Lidar Perception Evaluation}

We evaluate our perception by three different kinds of metrics. For each experiment, the data is collected by more than 5000 consecutive lidar frames in Town01. In Tab. \ref{table:LidarPerception}, $Recall$ indicates the truth positive detection of the surrounding vehicles by the LiDAR perception system. As the effective range decreases, the accuracy of the perception system becomes higher, thus the safety of the autonomous driving system is better. The 20-meter perception range setup is adapted according to the maximum speed and minimum braking distance. The $IoU$ represents the ratio of the intersection area between the detected and ground truth bounding box over the union area. The first two rows of overall object detection are relatively lower than the last two rows of moving object detection since we implement different strategies to align the bounding box to dynamic and static (speed estimated around zero) objects. From $E_{\theta}$, our optimized L-shape detection algorithm has a very high accuracy for estimating the object's orientation, with only a 2-3 degrees of estimation error. Speed estimation error $E_{v}$ is also under 0.6 $m/s$. When the perception system runs at 20 $Hz$, each time step is 0.05 second and the speed error is possibly caused by the LiDAR measurement noise. 

\begin{table}[H]
\caption{Evaluation for Lidar Perception}
\label{table:LidarPerception}
\begin{adjustbox}{width=8.5cm, center}
\begin{tabular}{ccccc}
\hline
\multicolumn{1}{l}{} & \multicolumn{2}{c}{Position} & Orientation (deg)         & Speed (m/s)          \\
Range                & $Recall$       & $mIoU$      & $E_{\theta}$  (Avg./Std.) & $E_{v}$ (Avg./ Std.) \\ \hline
$R$ = 20m            & 0.83           & 0.62        & 2.03 / 0.11               & 0.49 / 0.85          \\
$R$ = 15m            & 0.97           & 0.64        & 2.12 / 2.18               & 0.41 / 0.61          \\
Dyn. $R$ = 20m     & 0.94           & 0.71        & 3.08 / 6.05               & 0.59 / 2.23          \\
Dyn. $R$ = 15m     & 0.99           & 0.73        & 1.57 / 3.77               & 0.55 / 0.82          \\ \hline
\end{tabular}
\end{adjustbox}
\end{table}

\subsection{System Evaluation}
To evaluate the system performance by combing the lidar perception module with the motion planner, we ran the same simulations with the ground truth perception and recorded the platooning following distance and total execution time. The following distances are shown in the left-hand side of Fig.\ref{fig:fdistance}, and the time consumption of each step is also shown in Table.\ref{table:time consumption}. The overall time consumption is only increased by around 6 ms compared to the time consumption with ground truth information thanks to the asynchronous design for each node running in the perception framework.

\par The run-time table shows that lidar perception processing has an apparent slowdown compared to Town10 to Town01. The result is caused by the building and environmental features. Town01 is a country town which includes many little infrastructures like fences. It leads to over segments of vehicle candidate instancing as well as a rise in burden to the downstream processing like box alignment and tracking data association. But owing to the conservative policy design, the perception system can still achieve a zero-collision record in the 40 minutes long simulations.

\par
\begin{figure}[]{}
\centering
\begin{subfigure}[width=8cm]{0.7\textwidth}
    \includegraphics[scale = 0.2]{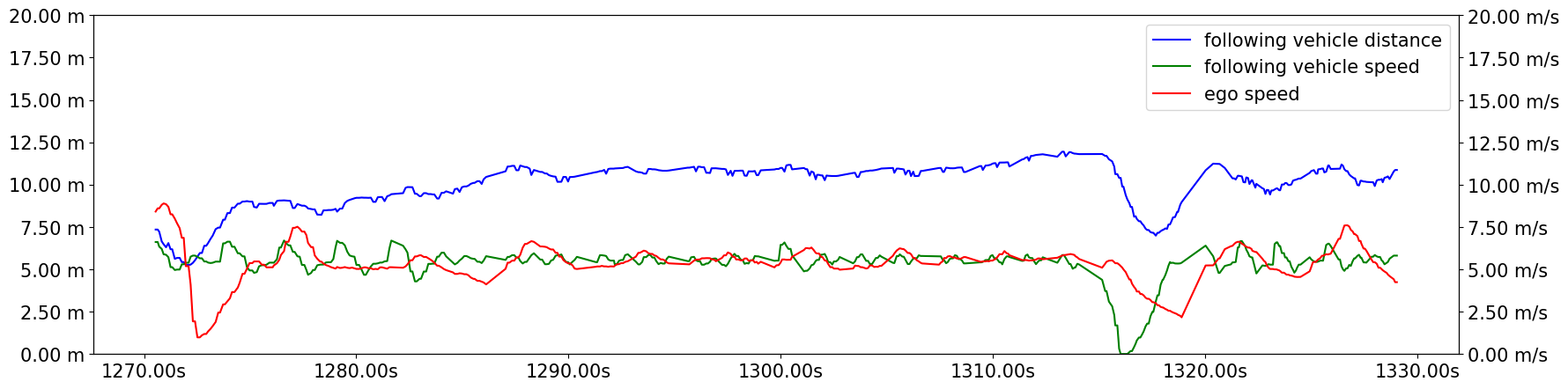}
\end{subfigure}
\begin{subfigure}[width=8cm]{0.7\textwidth}
    \includegraphics[scale = 0.2]{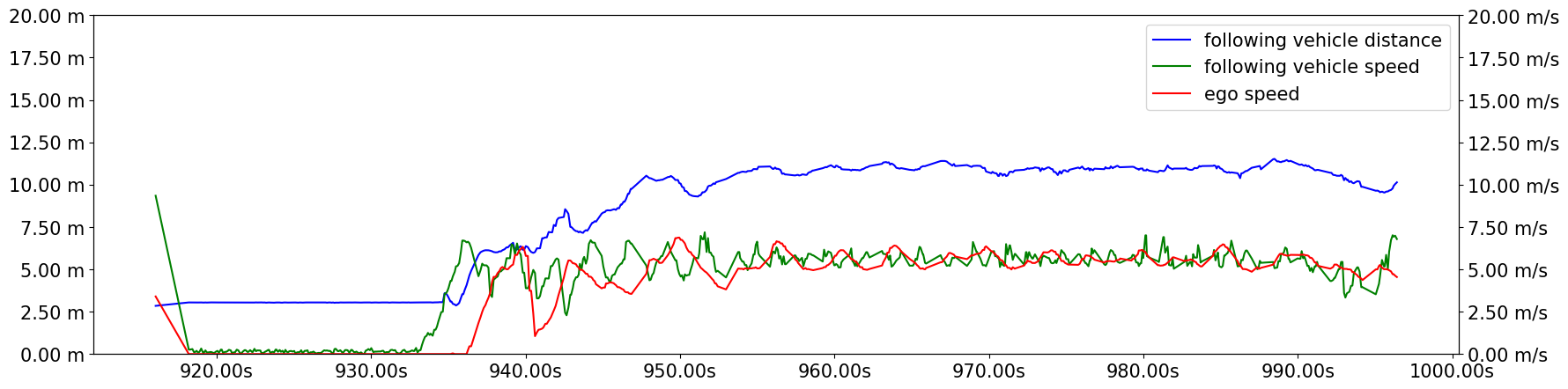}
\end{subfigure}
\caption{Top: Following distance data example by system using ground truth information. Bottom: Following distance data example by system with Lidar Perception.}
\label{fig:following_distance_example}
\end{figure}

 We also compared the platooning performance between different simulations within the same map. On the top of Fig. \ref{fig:following_distance_example} shows one example of the system with ground truth information. And the bottom of Fig. 
 \ref{fig:following_distance_example} shows one case with the Lidar perception. In both figures, the red, blue, and green lines represent the ego vehicle speed, the following vehicle speed, and the following distance. Compared two scenarios, we can see both systems performed a relative distance keeping as the speed increased. But the speed estimation from the Lidar perception module had more noise than the ground truth information, which caused variations in the distance keeping.

\section{Conclusion}
In this paper, we present an autonomous vehicle joint simulation system with both lidar perception and planning module deployed. Firstly, a traditional tracking-by-detection framework is employed to handle the lidar-based object detection. We optimize the L-shape detection algorithm and simplified the perception pipeline. Next, we introduce a new collision checking algorithm that allows relaxation of obstacle representations while also improves the long range collision check efficiency. We evaluate our system performance by simulating urban driving scenarios in CARLA simulator. Similar to the simulations using ground truth vehicle data, our system with LiDAR perception can run at an average of 20 Hz that meets the real-time requirement. Finally, within the range of 20 meters, the perception system gives accurate vehicle positions with a recall above 0.83 and mIOU above 0.62. Based on the dynamic velocity estimation, the average error of orientation is less than 3.08 degree and the average speed estimation error is less than 0.59 m/s. 










\bibliographystyle{IEEEtran}
\bibliography{root.bib}  
\end{document}